\pdfoutput=1

\documentclass[11pt]{article}

\usepackage{acl}
\usepackage{times}
\usepackage{latexsym}
\usepackage{booktabs}
\usepackage{multirow}

\usepackage[T1]{fontenc}

\usepackage[utf8]{inputenc}

\usepackage{microtype}

\usepackage{inconsolata}

\usepackage{graphicx}
\usepackage{tikz}
\usepackage{varwidth}
\usetikzlibrary{decorations.pathreplacing,calc}

\title{%
  Dream at SemEval-2026 Task 13:\\SALSA for Single-Pass Machine-Generated Code Detection%
}

\author{
  Ruslan Berdichevsky \\
  Dream Security Ltd. \\
  Tel Aviv, Israel \\
  \texttt{ruslan@dreamgroup.com}
  \And
  Shai Nahum-Gefen \\
  Dream Security Ltd. \\
  Tel Aviv, Israel \\
  \texttt{shai@dreamgroup.com}
  \And
  Elad Ben-Zaken \\
  Dream Security Ltd. \\
  Tel Aviv, Israel \\
  \texttt{elad@dreamgroup.com}
}

\begin{document}
\maketitle


\begin{abstract} Large language models have transformed code generation, raising concerns around authorship, assessment integrity, and software trust. SemEval-2026 Task~13 Subtask~A operationalizes detection as binary classification over code snippets, with a particular emphasis on out-of-distribution (OOD) generalization across unseen programming languages and application domains. We propose a SALSA-style formulation, Single-pass Autoregressive LLM Structured Classification, that maps each class to a dedicated output token and trains the model to emit a \emph{single-token} label in a structured response. Rather than engineering hand-crafted features or decision rules, this formulation \emph{delegates} the authorship decision to the model. To improve OOD robustness, we combine balanced sampling across languages with parameter-efficient fine-tuning and conservative training (low learning rate, single epoch) to avoid overfitting to the training domain. Our best system achieves OOD $F_1 = 0.789$ on the official leaderboard, substantially outperforming the CodeBERT baseline ($F_1 = 0.305$). 
\end{abstract}

\section{Introduction}
\begin{figure*}[t]
\centering
\includegraphics[width=\textwidth]{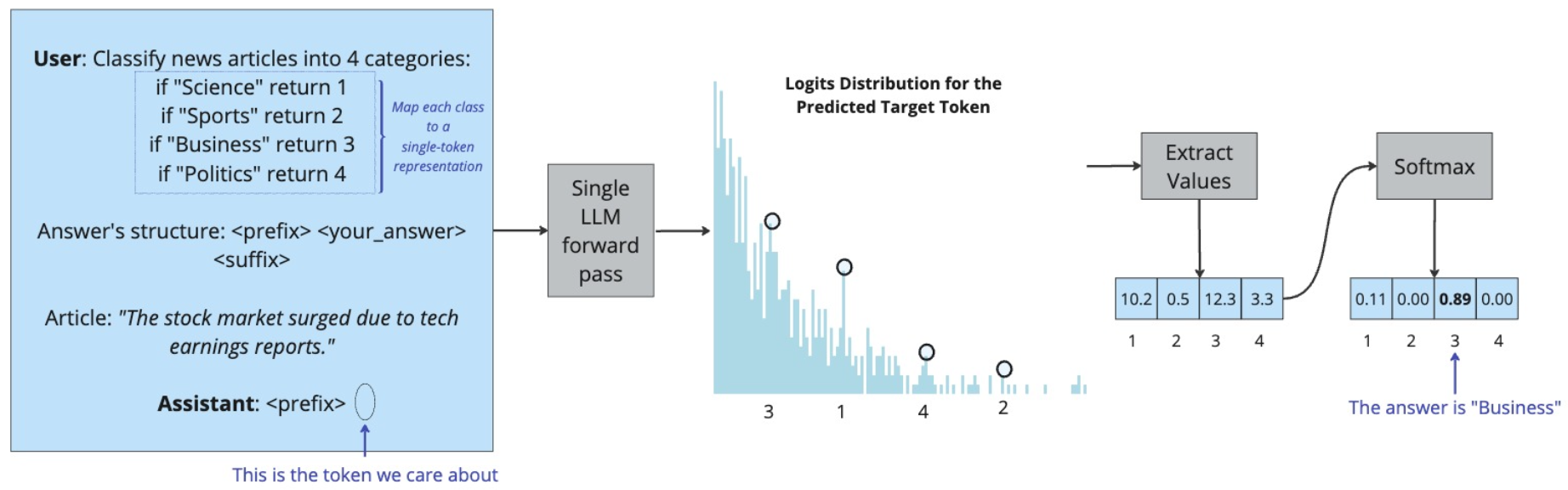}
\caption{Overview of the SALSA classification pipeline: the structured prompt maps each class to a single token; a single LLM forward pass produces logits for the target token position; the class-token logits are extracted and passed through a softmax to obtain the predicted label.}
\label{fig:salsa_overview}
\end{figure*}
Large language models (LLMs) have revolutionized code generation, but this has important consequences for programming skills, ethics, and assessment integrity, making the detection of LLM-generated code essential for maintaining accountability and standards \cite{orel-etal-2026-semeval-2026}.
While there is prior work on machine-generated code detection, it often lacks domain coverage and robustness, and typically supports only a small number of programming languages \cite{orel2025droid,orel-etal-2025-codet,orel2026aicdbenchchallengingbenchmark}.

SemEval-2026 Task~13 Subtask~A provides an important testbed by requiring a binary label indicating whether a code snippet is machine-generated.
A central challenge is generalization to out-of-distribution (OOD) settings, since practical detectors must handle unseen scenarios beyond the training distribution.

Rather than solving the problem via hand-designed features or decision 
rules, we \emph{delegate} the authorship decision to the model by 
specifying the task as a natural language instruction, since language 
models are strong general-purpose learners when prompted with task 
descriptions \cite{brown2020language}. However, using LLMs as classifiers 
this way is often brittle: the model may emit verbose explanations, 
inconsistent label formats, or multi-token labels. SALSA\footnote{Our code is publicly available at \url{https://github.com/dreamgroupai-ai/SALSA}.}, Single-pass Autoregressive LLM Structured Classification~\citep{berdichevsky2025salsa}, addresses this
by mapping each class to a distinct output token and structuring the
prompt with clear delimiters and a direct authorship question, turning
classification into a reliable constrained next-token prediction.

During fine-tuning, the model simultaneously learns task alignment (following the classification instruction) and in-domain expertise (adapting to the training distribution).
Since our primary goal is out-of-distribution (OOD) performance, we bias training toward preserving task alignment and general capabilities rather than over-fitting to the source domain.
To this end, we run only one epoch and select the best checkpoint based on validation; prior work suggests that earlier training tends to improve general-purpose features, while later stages may over-specialize to the source domain and degrade OOD performance \cite{liu2024early}.
We also use a low learning rate, as we empirically observe that larger learning rates tend to reduce OOD performance while improving in-distribution metrics.
Together, these choices keep the fine-tuned model closer to the base model, which tends to improve generalization.

SALSA has shown consistent results across diverse classification settings, including the GLUE benchmark \cite{wang2019glue} and content moderation \cite{jigsaw-agile-community-rules,kaggle_jigsaw_salsa_writeup}; here we apply it to machine-generated code detection and report strong OOD performance ($F_1=0.789$).

\section{Background}
Subtask~A is a binary classification setting: given an input code snippet, the system predicts \texttt{0} (human-written) or \texttt{1} (machine-generated).
The task is designed to stress-test generalization across both programming languages and application scenarios: training data uses \texttt{C++}, \texttt{Python}, and \texttt{Java} from an algorithmic domain, while evaluation includes (i) seen vs.~unseen languages and (ii) seen vs.~unseen domains (research/production), yielding four settings in total.

We use the official dataset distribution released by the task organizers.
The dataset provides the code snippet (\texttt{code}), the binary label (\texttt{label}), and metadata including programming language (\texttt{language}).
The released sizes for Subtask~A are 500K training samples (238K human-written and 262K machine-generated) and 100K validation samples, and a private test set of 500K samples for which no labels or metadata are provided.

\section{System Overview}
\label{sec:system}
\subsection{SALSA}
\begin{figure*}[t]
\begin{tikzpicture}

\node[anchor=north east, inner sep=0pt] (box) at (12,0) {%
\fbox{%
\begin{varwidth}{0.85\textwidth}
\ttfamily
\footnotesize
\noindent
\textless{}|begin\_of\_text|\textgreater\\[2pt]
\textless{}|im\_start|\textgreater{} user /no\_think\\[2pt]
Given the code: \textless CODE\textgreater\\[2pt]
\hspace*{1em}s = input()\\[2pt]
\hspace*{1em}pal = s + s[::-1]\\[2pt]
\hspace*{1em}print(pal)\\[2pt]
\textless/ CODE\textgreater\\[2pt]
is the code machine generated?\\[2pt]
provide answer in format:\\[2pt]
\textless ANSWER\textgreater\#Number\textless/ANSWER\textgreater\\[2pt]
where the number is one of the following:\\[2pt]
\hspace*{1em}0 - No\\[2pt]
\hspace*{1em}1 - Yes\\[2pt]
\textless{}|im\_end|\textgreater\\[2pt]
\textless{}|im\_start|\textgreater{} assistant\\[2pt]
\textless think\textgreater\ \textless/think\textgreater\\[2pt]
\textless ANSWER\textgreater \#\textless/ANSWER\textgreater\\[2pt]
\textless{}|im\_end|\textgreater
\end{varwidth}
}};

\path (box.north west) coordinate (NW);

\draw [decorate,decoration={brace,amplitude=4pt,mirror}, thick]
  ($(NW)+(0,-0.15)$) -- ($(NW)+(0,-0.55)$)
  node[midway,xshift=-1.5cm]{\textbf{BOS token}};

\draw [decorate,decoration={brace,amplitude=4pt,mirror}, thick]
  ($(NW)+(0,-0.65)$) -- ($(NW)+(0,-4.35)$)
  node[midway,xshift=-1.3cm]{\textbf{Task + input}};

\draw [decorate,decoration={brace,amplitude=4pt,mirror}, thick]
  ($(NW)+(0,-4.5)$) -- ($(NW)+(0,-5.45)$)
  node[midway,xshift=-1.5cm]{\textbf{Class mapping}};

\draw [decorate,decoration={brace,amplitude=4pt,mirror}, thick]
  ($(NW)+(0,-5.6)$) -- ($(NW)+(0,-7.1)$)
  node[midway,xshift=-1.6cm]{\begin{tabular}{l}\textbf{Answer format} \\ with masked token\end{tabular}};

\end{tikzpicture}
\caption{Example SALSA-structured prompt for machine-generated code detection (Subtask~A).}
\label{fig:prompt_construction}
\end{figure*}
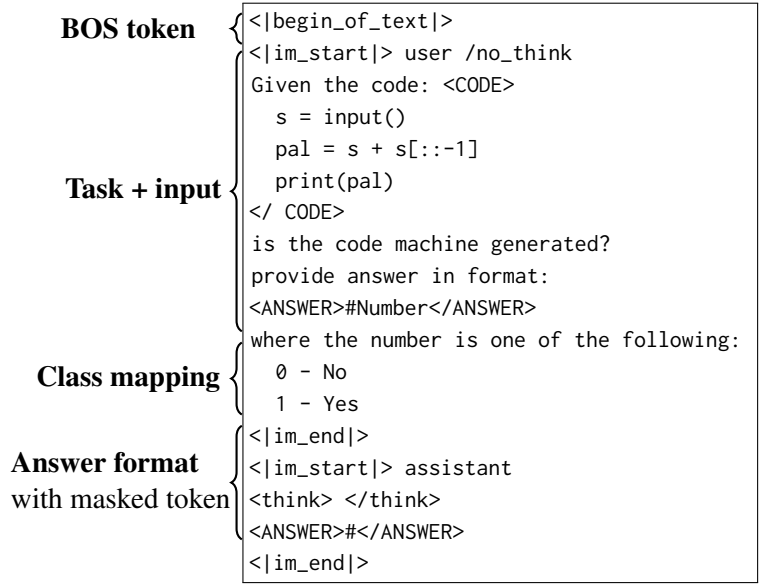

SALSA reframes classification as single-token generation over an autoregressive LLM.
Each class is mapped to a unique output token, and the prompt is structured to elicit exactly that token at a fixed position.
At inference time, a single forward pass suffices: the output logits are projected onto the class tokens only, yielding an efficient classifier with minimal decoding overhead.
Figure~\ref{fig:salsa_overview} illustrates the end-to-end pipeline, and Figure~\ref{fig:prompt_construction} shows an example prompt for our task.

\paragraph{Class-to-token mapping.}
For Subtask~A we use label tokens \texttt{"0"} and \texttt{"1"} (a single token each under the model tokenizer).

\paragraph{Structured prompting.}
We wrap the code snippet in delimiters and ask the model whether the code is machine generated.
We require the answer in the form \texttt{<ANSWER>\#</ANSWER>} where \texttt{\#} is \texttt{0} or \texttt{1}.
Figure~\ref{fig:prompt_construction} shows the prompt structure used for the task. The \texttt{/no\_think} flag and an empty \texttt{<think>} block suppress chain-of-thought reasoning in Qwen3, keeping the output to a single label token \cite{qwen3}.

\subsection{Implementation Details}
\paragraph{Training objective}
We implement SALSA by supervising only the label token position in the assistant response.
Concretely, the model is trained with a causal language modeling objective where the loss is evaluated exclusively at the final occurrence of the label token, so gradients update the model primarily to improve the next-token distribution over the target classes.

\paragraph{Parameter-efficient fine-tuning}
We fine-tune with LoRA, Low-Rank Adaptation \cite{hu2022lora}, to reduce overfitting risk and speed up training.

\paragraph{Conservative fine-tuning.}
We use a low learning rate ($5{\times}10^{-6}$), train for one epoch, and select the best checkpoint on validation.
We empirically observe that more aggressive training reduces OOD performance while improving in-distribution metrics.
In particular, training beyond a single epoch consistently degrades OOD generalization, suggesting that extended exposure to the source domain causes the model to over-specialize at the expense of its pretrained representations.

\paragraph{Dataset splits.}
Table~\ref{tab:data_dist} shows the per-language and per-label breakdown of the training (500K) and validation (100K) splits.
The dataset is heavily skewed toward \texttt{Python}, which accounts for 91.5\% of all samples, while \texttt{C++} and \texttt{Java} together comprise less than 9\%.
The label distribution is roughly balanced overall (47.7\% human vs.\ 52.3\% machine-generated), and this balance is preserved within each language; the validation split closely mirrors the training proportions.

\begin{table}[t]
\centering
\setlength{\tabcolsep}{3.5pt}
\renewcommand{\arraystretch}{0.92}
\footnotesize
\begin{tabular}{llrrr}
\toprule
\textbf{Split} & \textbf{Language} & \textbf{Human (0)} & \textbf{Machine (1)} & \textbf{Total} \\
\midrule
\multirow{4}{*}{Train}
  & C++       &  11{,}147 &  12{,}245 &  23{,}392 \\
  & Java      &   9{,}225 &  10{,}077 &  19{,}302 \\
  & Python    & 218{,}103 & 239{,}203 & 457{,}306 \\
  & \textbf{Total} & 238{,}475 & 261{,}525 & 500{,}000 \\
\midrule
\multirow{4}{*}{Val.}
  & C++       &  2{,}230 &  2{,}449 &  4{,}679 \\
  & Java      &  1{,}845 &  2{,}015 &  3{,}860 \\
  & Python    & 43{,}620 & 47{,}841 & 91{,}461 \\
  & \textbf{Total} & 47{,}695 & 52{,}305 & 100{,}000 \\
\bottomrule
\end{tabular}
\caption{Label distribution by split and programming language.}
\label{tab:data_dist}
\end{table}

\paragraph{Balanced sampling}
As shown in Table~\ref{tab:data_dist}, the training set is highly imbalanced across programming languages.
To prevent majority-language dominance, we balance the training data by sampling an equal number of examples from each \texttt{(language, label)} group.
The number of samples drawn per group is determined by the size of the smallest group; samples are drawn without repetition, so each training example is used at most once.
This discards a substantial portion of Python training data, a deliberate trade-off: preserving Python's dominance would steer fine-tuning toward in-domain expertise on Python-specific patterns at the expense of the task alignment carried over from the pretrained starting point.
In preliminary experiments, the unbalanced distribution yielded weaker task alignment (lower OOD $F_1$), consistent with this view.

\paragraph{Inference}
Our inference follows the SALSA principle of restricted logit projection:
we generate a single token and read the model's log-probabilities for the class tokens only.
The predicted label is the argmax over the two target token logits.
We achieve high-throughput inference by merging the LoRA adapter into the base model and using vLLM \cite{kwon2023efficient} to compute logprobs efficiently.

\section{Experimental Setup}
\paragraph{Configuration.}
We fine-tune Qwen3-8B, Qwen3-32B and Qwen2.5-72B-Instruct \cite{qwen2.5} with LoRA adapters using the SALSA 
structured prompt described in Section~\ref{sec:system}. 
Full hyperparameter details are provided in Appendix~\ref{sec:appendix}.
We train for up to 1 epoch, selecting the checkpoint with the best 
validation $F_1$, and use left truncation at 8192 tokens to
preserve the label token at the end of the prompt.

\paragraph{Metric.}
We report macro-$F_1$.
For in-distribution performance, we evaluate on the official Kaggle validation split as described in the challenge ($F_1=0.996$ in our experiment).
For out-of-distribution performance, we report the competition OOD test $F_1$ (0.789) on the Kaggle leaderboard.

\section{Results}
Table~\ref{tab:results} summarizes our results.
The official CodeBERT baseline scores 0.305 OOD $F_1$.
Our best submission, Qwen2.5-72B-Instruct fine-tuned with SALSA, achieves 0.789 OOD $F_1$, a substantial improvement over both the baseline and the zero-shot upper bound for that model (0.449).

\begin{table}[t]
\centering
\footnotesize
\setlength{\tabcolsep}{2pt}
\begin{tabular}{@{}lccc@{}}
\toprule
\textbf{Model} & \textbf{Test $F_1$} & \textbf{Val.\ $F_1$} & \textbf{Train $F_1$} \\
\midrule
\multicolumn{4}{l}{\textit{Baseline}} \\
CodeBERT & 0.305 & -- & -- \\
\midrule
\multicolumn{4}{l}{\textit{SALSA Zero-shot - without tuning}} \\
Qwen3-Next-80B-A3B-Instruct   & 0.591 & 0.672 & 0.662 \\
Qwen2.5-72B-Instruct          & 0.449 & 0.359 & 0.358 \\
Qwen3-32B                     & 0.508 & 0.530 & 0.528 \\
Qwen3-8B                      & 0.436 & 0.523 & 0.523 \\
\midrule
\multicolumn{4}{l}{\textit{SALSA Tuned}} \\
Qwen2.5-72B-Instruct          & 0.789 & 0.996 & 0.996 \\
Qwen3-32B                     & 0.760 & 0.994 & 0.994 \\
Qwen3-32B official$^\dagger$  & 0.730 & 0.989 & 0.982 \\
Qwen3-8B                      & 0.450 & 0.991 & 0.991 \\
\bottomrule
\end{tabular}
\caption{Results on SemEval-2026 Task~13 Subtask~A. CodeBERT is the organizers' baseline. Test $F_1$ denotes the leaderboard OOD score. Zero-shot uses the SALSA prompt without fine-tuning. $^\dagger$\emph{Note:} The official Kaggle submission achieved a test (OOD) $F_1$ of 0.73 due to suboptimal checkpoint selection; a post-competition ablation improved this to 0.76.}
\label{tab:results}
\end{table}

\paragraph{Zero-shot prediction bias.}
The precision/recall breakdown in Table~\ref{tab:zeroshot_bias} reveals that the aggregate $F_1$ scores in Table~\ref{tab:results} mask qualitatively different failure modes.
Qwen3-Next-80B-A3B-Instruct is the only zero-shot model with balanced precision and recall (0.707 / 0.638), while Qwen3-32B and Qwen2.5-72B-Instruct are strongly biased toward predicting human-written code, achieving recall of only 0.242 and 0.035 respectively despite high precision.
This near-total failure to detect machine-generated code explains the val/OOD inversion for Qwen2.5-72B-Instruct (val $F_1$=0.359 vs.\ OOD $F_1$=0.449): the OOD distribution is apparently less dominated by machine-generated examples, so the model's conservative bias is less penalized.
The breakdown also directly motivates our balanced sampling strategy: a model that defaults to predicting ``human'' will score well on a skewed distribution but generalize poorly.
The per-language breakdown in Appendix~\ref{sec:zeroshot_detail} shows no strong language-level skew in the validation set.

\begin{table}[t]
\centering
\footnotesize
\setlength{\tabcolsep}{3pt}
\begin{tabular}{lccc}
\toprule
\textbf{Model (Zero-shot)} & \textbf{Prec.} & \textbf{Rec.} & \textbf{Val $F_1$} \\
\midrule
Qwen3-Next-80B-A3B-Instruct & 0.707 & 0.638 & 0.672 \\
Qwen3-32B                   & 0.844 & 0.242 & 0.530 \\
Qwen3-8B                    & 0.572 & 0.397 & 0.523 \\
Qwen2.5-72B-Instruct        & 0.910 & 0.035 & 0.359 \\
\bottomrule
\end{tabular}
\caption{Zero-shot validation performance. Prec./Rec. denote precision and recall for the machine-generated class; $F_1$ is macro-averaged. The high-precision/low-recall asymmetry indicates a bias toward predicting human-written code; per-language breakdown in Appendix~\ref{sec:zeroshot_detail}.}
\label{tab:zeroshot_bias}
\end{table}

\paragraph{Validation results.}
All fine-tuned models achieve near-perfect validation $F_1$: 0.991 (8B), 0.994 (32B), and 0.996 (72B), essentially matching their training-split $F_1$.
The absence of a train--val gap indicates in-distribution fitting without overfitting.
While the differences are small in absolute terms, the scores increase monotonically with model size, consistent with the OOD trend.
We use the validation score for checkpoint selection within training, but do not draw strong conclusions from cross-model comparisons given the narrow range. Extended validation error analysis is provided in Appendix~\ref{sec:val_error_analysis}.

\paragraph{Effect of fine-tuning.}
Notably, all zero-shot models already outperform the CodeBERT baseline (0.305), with scores of 0.359--0.672, suggesting that general-purpose LLMs carry useful prior knowledge for this task even without task-specific training.
Fine-tuning with SALSA yields substantial further gains for larger models, with both Qwen2.5-72B-Instruct and Qwen3-32B improving markedly over their zero-shot counterparts. For Qwen3-8B, the gain is negligible, indicating that fine-tuning alone is insufficient for small models in this setting.

\section{Conclusion}
We applied SALSA to SemEval-2026 Task~13 Subtask~A, achieving an OOD $F_1$ of 0.789 with Qwen2.5-72B-Instruct, placing among the top scores on the official leaderboard.
SALSA is a natural fit for this task: by reducing the output space to two tokens and delegating the authorship decision to the model via structured prompting and tuning, it yields stable, efficient classification.
OOD performance improves monotonically with model scale, which we attribute to larger models retaining more of their pretrained knowledge after conservative fine-tuning rather than over-specializing to the training domain.
Interestingly, Qwen2.5-72B-Instruct leads after fine-tuning, yet its zero-shot performance is clearly weaker than Qwen3 models, suggesting that Qwen3's stronger instruction alignment is beneficial in the zero-shot regime while Qwen2.5's broader pretraining knowledge transfers better under fine-tuning. The MoE model achieves the strongest zero-shot alignment; since our current implementation does not support MoE fine-tuning, exploring that direction is left for future work.
Future work should focus on improving task instruction alignment without inducing training-domain specialization.

\bibliography{references}

\appendix

\begin{table*}[t]
\centering
\small
\setlength{\tabcolsep}{4pt}
\begin{tabular}{llrrrrrrrrr}
\toprule
\textbf{Model} & \textbf{Lang} & \textbf{N} & \textbf{TP} & \textbf{TN} & \textbf{FP} & \textbf{FN} & \textbf{$F_1$} & \textbf{Prec.} & \textbf{Rec.} \\
\midrule
\multirow{4}{*}{Qwen3-Next-80B-A3B-Instruct}
  & C++    &  4{,}679 & 1{,}277 & 2{,}033 &   197 & 1{,}172 & 0.700 & 0.866 & 0.521 \\
  & Java   &  3{,}860 & 1{,}040 & 1{,}710 &   135 &   975   & 0.704 & 0.885 & 0.516 \\
  & Python & 91{,}461 & 31{,}028 & 30{,}134 & 13{,}486 & 16{,}813 & 0.669 & 0.697 & 0.649 \\
  & \textbf{All} & \textbf{100{,}000} & \textbf{33{,}345} & \textbf{33{,}877} & \textbf{13{,}818} & \textbf{18{,}960} & \textbf{0.672} & \textbf{0.707} & \textbf{0.638} \\
\midrule
\multirow{4}{*}{Qwen3-32B}
  & C++    &  4{,}679 &   564 & 2{,}174 &    56 & 1{,}885 & 0.530 & 0.910 & 0.230 \\
  & Java   &  3{,}860 &   361 & 1{,}808 &    37 & 1{,}654 & 0.490 & 0.907 & 0.179 \\
  & Python & 91{,}461 & 11{,}746 & 41{,}368 & 2{,}252 & 36{,}095 & 0.532 & 0.839 & 0.246 \\
  & \textbf{All} & \textbf{100{,}000} & \textbf{12{,}671} & \textbf{45{,}350} & \textbf{2{,}345} & \textbf{39{,}634} & \textbf{0.530} & \textbf{0.844} & \textbf{0.242} \\
\midrule
\multirow{4}{*}{Qwen3-8B}
  & C++    &  4{,}679 &   875 & 1{,}318 &   912 & 1{,}574 & 0.464 & 0.490 & 0.357 \\
  & Java   &  3{,}860 &   666 & 1{,}274 &   571 & 1{,}349 & 0.490 & 0.538 & 0.331 \\
  & Python & 91{,}461 & 19{,}213 & 29{,}592 & 14{,}028 & 28{,}628 & 0.528 & 0.578 & 0.402 \\
  & \textbf{All} & \textbf{100{,}000} & \textbf{20{,}754} & \textbf{32{,}184} & \textbf{15{,}511} & \textbf{31{,}551} & \textbf{0.523} & \textbf{0.572} & \textbf{0.397} \\
\midrule
\multirow{4}{*}{Qwen2.5-72B-Instruct}
  & C++    &  4{,}679 &   106 & 2{,}188 &    42 & 2{,}343 & 0.364 & 0.716 & 0.043 \\
  & Java   &  3{,}860 &    60 & 1{,}815 &    30 & 1{,}955 & 0.352 & 0.667 & 0.030 \\
  & Python & 91{,}461 &  1{,}644 & 43{,}512 &   108 & 46{,}197 & 0.360 & 0.938 & 0.034 \\
  & \textbf{All} & \textbf{100{,}000} & \textbf{1{,}810} & \textbf{47{,}515} & \textbf{180} & \textbf{50{,}495} & \textbf{0.359} & \textbf{0.910} & \textbf{0.035} \\
\bottomrule
\end{tabular}
\caption{Zero-shot per-language results on the validation set. TP/TN/FP/FN are counts for the machine-generated class (label=1). The human-prediction bias of Qwen2.5-72B-Instruct and Qwen3-32B is consistent across all languages.}
\label{tab:zeroshot_lang}
\end{table*}

\section{Hardware and Hyperparameters}
\label{sec:appendix}

All models were fine-tuned on 2 NVIDIA B200 GPUs using DeepSpeed~\cite{rasley2020deepspeed} ZeRO stage~0 and for more than 70B stage~3.
We trained for at most one epoch with a learning rate of $5{\times}10^{-6}$, a per-GPU batch size of~1, and gradient accumulation over 32 steps (effective batch size~64).
Input sequences were left-truncated to 8{,}192 tokens to ensure the label token is always retained at the end of the prompt.
LoRA adapters were configured with rank $r{=}16$, scaling factor $\alpha{=}32$, and dropout~$0.05$.

\section{Inference Runtime}
\label{sec:runtime}

Table~\ref{tab:runtime} reports wall-clock inference times for a full evaluation
run over the 500K test records, measured on 2 B200 GPUs using vLLM \cite{kwon2023efficient}.

\begin{table}[h]
\centering
\small
\begin{tabular}{lc}
\toprule
\textbf{Model} & \textbf{Runtime (hh:mm)} \\
\midrule
Qwen3-Next-80B-A3B-Instruct & 1:59 \\
Qwen2.5-72B-Instruct & 4:10 \\
Qwen3-32B            & 2:11 \\
Qwen3-8B             & 0:56 \\
\bottomrule
\end{tabular}
\caption{Wall-clock inference time for 500K records on 2 B200 GPUs using vLLM.}
\label{tab:runtime}
\end{table}

\section{Zero-Shot Per-Language Breakdown}
\label{sec:zeroshot_detail}

Table~\ref{tab:zeroshot_lang} reports precision, recall, and macro-$F_1$ for each zero-shot model broken down by programming language on the validation set.
The human-prediction bias of Qwen2.5-72B-Instruct and Qwen3-32B is consistent across all three languages, confirming it is a model-level property rather than a language-specific artifact.
Qwen3-Next-80B-A3B-Instruct maintains the most balanced recall across languages, though recall drops notably for \texttt{C++} and \texttt{Java} compared to \texttt{Python}.

\section{Validation Error Analysis}
\label{sec:val_error_analysis}
 
We analyze the prediction errors of our lead Qwen3-32B checkpoint on the 100K validation split.
The model reaches a macro-$F_1$ of $0.994$, with 601 misclassified samples.
 
\paragraph{Confusion Matrix.}
A slight bias toward the human label can be observed in Table~\ref{tab:val_confusion}. However, because the overall separation is nearly perfect, this effect is too small to support a strong conclusion.

\begin{table}
\centering
\small
\begin{tabular}{lrr}
\toprule
 & \textbf{Pred.\ Human (0)} & \textbf{Pred.\ AI (1)} \\
\midrule
\textbf{Actual Human (0)} & 47{,}496 & 199 \\
\textbf{Actual AI (1)}    &      402 & 51{,}903 \\
\bottomrule
\end{tabular}
\caption{Validation confusion matrix for Qwen3-32B.}
\label{tab:val_confusion}
\end{table}

\paragraph{Per-language breakdown.}
Despite balanced sampling across (\texttt{language}, \texttt{label}) groups, \texttt{Python} retains a noticeably lower error rate than \texttt{C++} and \texttt{Java} (Table~\ref{tab:val_per_lang}), likely reflecting broader Python coverage in Qwen3's pretraining.
 
\begin{table}
\centering
\small
\begin{tabular}{lrrccc}
\toprule
\textbf{Lang.} & \textbf{N} & \textbf{Err.} & \textbf{Prec.} & \textbf{Rec.} & \textbf{$F_1$} \\
\midrule
Python & 91{,}461 & 462 & 0.9949 & 0.9950 & 0.9949 \\
C++    &  4{,}679 &  71 & 0.9845 & 0.9853 & 0.9848 \\
Java   &  3{,}860 &  68 & 0.9821 & 0.9828 & 0.9824 \\
\bottomrule
\end{tabular}
\caption{Per-language validation performance for fine-tuned Qwen3-32B. ``Err.'' is the number of misclassified samples; precision, recall, and $F_1$ are macro-averaged.}
\label{tab:val_per_lang}
\end{table}

\paragraph{Per-generator error rate.}
Table~\ref{tab:val_per_gen} reports the validation error rate for each AI source model.
The hardest generators to detect are from the IBM Granite family, followed by \texttt{CodeLlama-34b-Instruct} and \texttt{deepseek-coder-6.7b-base}.
The table suggests that \emph{base} variants are often harder to detect than \emph{instruct} variants, though this trend is not uniform.
This may indicate that instruction tuning can produce more stylized, template-like outputs that are easier to separate from human code.
 
\begin{table}[!t]
\centering
\small
\setlength{\tabcolsep}{4pt}
\renewcommand{\arraystretch}{1.0}
\begin{tabular*}{\columnwidth}{@{\extracolsep{\fill}}lrr@{}}
\toprule
\textbf{Generator} & \textbf{N} & \textbf{Err.\ rate} \\
\midrule
Yi-Coder-9B-Chat                  & 1{,}665  & 0.00\% \\
Phi-3-mini-4k-instruct            & 1{,}667  & 0.00\% \\
Phi-3-small-8k-instruct           & 1{,}814  & 0.06\% \\
deepseek-coder-1.3b-base          & 1{,}113  & 0.09\% \\
Phi-3-medium-4k-instruct          & 3{,}165  & 0.13\% \\
deepseek-coder-1.3b-instruct      &    682   & 0.15\% \\
starcoder2-7b                     & 1{,}350  & 0.15\% \\
Qwen2.5-Coder-1.5B-Instruct       & 1{,}903  & 0.16\% \\
starcoder2-3b                     & 1{,}760  & 0.17\% \\
Yi-Coder-1.5B                     & 1{,}732  & 0.17\% \\
CodeLlama-70b-Instruct-hf         & 1{,}742  & 0.23\% \\
Phi-3.5-mini-instruct             & 2{,}047  & 0.24\% \\
Llama-3.2-1B                      &    760   & 0.26\% \\
Yi-Coder-1.5B-Chat                & 1{,}341  & 0.30\% \\
Llama-3.2-3B                      & 1{,}550  & 0.32\% \\
deepseek-coder-6.7b-instruct      & 1{,}126  & 0.36\% \\
Qwen2.5-Coder-7B-Instruct         & 1{,}324  & 0.38\% \\
\emph{human}                      & 47{,}695 & 0.42\% \\
Qwen2.5-Coder-32B-Instruct        & 1{,}597  & 0.50\% \\
CodeLlama-7b-hf                   & 1{,}128  & 0.53\% \\
Qwen2.5-Coder-1.5B                & 1{,}306  & 0.54\% \\
codegemma-2b                      & 1{,}484  & 0.67\% \\
Llama-3.1-8B                      & 1{,}351  & 0.74\% \\
Llama-3.1-8B-Instruct             & 1{,}653  & 0.85\% \\
phi-2                             & 1{,}966  & 0.97\% \\
starcoder2-15b                    & 1{,}872  & 1.07\% \\
starcoder                         & 1{,}857  & 1.24\% \\
Qwen2.5-Coder-7B                  & 1{,}427  & 1.33\% \\
codegemma-7b                      & 1{,}258  & 1.43\% \\
Yi-Coder-9B                       & 1{,}950  & 1.49\% \\
Llama-3.3-70B-Instruct            & 1{,}756  & 1.59\% \\
deepseek-coder-6.7b-base          & 1{,}207  & 2.15\% \\
CodeLlama-34b-Instruct-hf         & 1{,}544  & 2.59\% \\
granite-8b-code-instruct-4k       &    846   & 2.60\% \\
granite-8b-code-base-4k           & 1{,}362  & 4.11\% \\
\bottomrule
\end{tabular*}
\caption{Per-generator validation error rate for fine-tuned Qwen3-32B.}
\label{tab:val_per_gen}
\end{table}

\end{document}